\documentclass[10pt, journal]{IEEEtran}
\usepackage{courier}
\usepackage{times}
\usepackage{soul}
\usepackage{url}
\usepackage{array}
\usepackage[utf8]{inputenc}
\usepackage{graphicx}
\usepackage{amsmath}
\usepackage{booktabs}
\usepackage{algorithm}
\usepackage{algorithmic}
\usepackage[switch]{lineno}
\usepackage{multirow}
\usepackage{multicol}
\usepackage{graphicx}
\usepackage{subfigure}
\usepackage{graphicx}
\usepackage{amssymb}
\usepackage{xcolor}
\usepackage{url} 
\usepackage[colorlinks=false, pdfborder={0 0 0}]{hyperref}

\let\oldurl\url
\renewcommand{\url}[1]{\textcolor{green!50!black}{\oldurl{#1}}}

\newcommand{\myref}[1]{Eq.(\ref{#1})}
\newcommand{\RED}[1]{\textcolor{black}{#1}}

 \makeatletter
    \newcommand{\thickhline}{%
        \noalign {\ifnum 0=`}\fi \hrule height 1pt
        \futurelet \reserved@a \@xhline
    }
    \newcolumntype{"}{@{\vrule width 1pt}}
    \makeatother
\makeatletter
\let\ftype@table\ftype@figure
\makeatother

\let\oldFootnote\footnote
\newcommand\nextToken\relax

\renewcommand\footnote[1]{%
    \oldFootnote{#1}\futurelet\nextToken\isFootnote}

\newcommand\isFootnote{%
    \ifx\footnote\nextToken\textsuperscript{,}\fi}

\begin{document}

\title{Data-Free Class-Incremental Gesture Recognition with Prototype-Guided Pseudo Feature Replay}

\author{Hongsong~Wang, Ao~Sun, Jie~Gui, and~Liang~Wang, Fellow, IEEE % <-this % stops a space
\IEEEcompsocitemizethanks{
% \IEEEcompsocthanksitem This work was supported by National Science Foundation of China (62302093, 62172090, 52441503), Jiangsu Province Natural Science Fund (BK20230833), the Fundamental Research Funds for the Central Universities (2242025K30024) and the Open Research Fund of the State Key Laboratory of Multimodal Artificial Intelligence Systems (E5SP060116). We thank the Big Data Computing Center of Southeast University for providing the facility support on the numerical calculations.
\IEEEcompsocthanksitem H.~Wang is with School of Computer Science and Engineering, Key Laboratory of New Generation Artificial Intelligence Technology and Its Interdisciplinary Applications, Ministry of Education, Southeast University, Nanjing 210096, China (hongsongwang@seu.edu.cn).
\IEEEcompsocthanksitem A.~Sun and J.~Gui are with School of Cyber Science and Engineering, Engineering Research Center of Blockchain Application, Supervision And Management (Southeast University), Ministry of Education, Southeast University, Nanjing 210096, China (\{aosun1022,guijie\}@seu.edu.cn).
\IEEEcompsocthanksitem L. Wang is with New Laboratory of Pattern Recognition (NLPR), State Key Laboratory of Multimodal Artificial Intelligence Systems (MAIS), Institute of Automation, Chinese Academy of Sciences (CASIA), and also with School of Artificial Intelligence, University of Chinese Academy of Sciences (wangliang@nlpr.ia.ac.cn).
%  (e-mail: yhuang@nlpr.ia.ac.cn; wangliang@nlpr.ia.ac.cn)
 }% <-this % stops an unwanted space
% \thanks{Manuscript received April 19, 2005; revised August 26, 2015.}
}

\markboth{JOURNAL OF LATEX CLASS FILES,~Vol.~xx, No.~xx, xx~2017}%
{Shell \MakeLowercase{\textit{et al.}}: Bare Demo of IEEEtran.cls for IEEE Journals}

\maketitle

\begin{abstract}
Gesture recognition is an important research area in the field of computer vision. Most existing efforts focus on close-set scenarios, thereby limiting the capacity to effectively handle unseen or novel gestures. We aim to address class-incremental gesture recognition, which entails the ability to accommodate new and previously unseen gestures over time. Specifically, we introduce a Prototype-Guided Pseudo Feature Replay framework for data-free class-incremental learning. This framework comprises four components: Pseudo Feature Generation with Batch Prototypes (PFGBP), Variational Prototype Replay for old classes, Truncated Cross-Entropy for new classes, and Continual Classifier Re-Training. To tackle the issue of catastrophic forgetting, the PFGBP dynamically generates a diversity of pseudo features in an online manner, leveraging class prototypes of old classes along with batch class prototypes of new classes. Furthermore, the Variational Prototype Replay enforces consistency between the classifier's weights and the prototypes of old classes, leveraging class prototypes and covariance matrices to enhance robustness and generalization capabilities. The Truncated Cross-Entropy mitigates the impact of domain differences of the classifier caused by pseudo features. Finally, the Continual Classifier Re-Training training strategy is designed to prevent overfitting to new classes and ensure the stability of features extracted from old classes. Extensive experiments conducted on two widely used gesture recognition datasets, namely SHREC 2017 3D and EgoGesture 3D, demonstrate that our approach outperforms existing state-of-the-art methods by 11.8\% and 12.8\% in terms of mean global accuracy, respectively. The code is available on \url{https://github.com/sunao-101/PGPFR-3/}.
\end{abstract}

\maketitle
\IEEEdisplaynontitleabstractindextext
\IEEEpeerreviewmaketitle

\section{Introduction} \label{sec:intro}
% cross-modal retrieval~\cite{li2021memorize,huang2018bi,xie2020multi}, 
% It serves as the basis for various visual and language tasks
% text-to-image synthesis~\cite{xu2018attngan,tao2020df},  liu2017pku-mmd,
% 1) gesture recognition
\IEEEPARstart{H}{and} gesture recognition is an essential research problem with broad applications in fields such as human-computer interaction, virtual reality, and assistive technology. Recently, data-driven gesture recognition methods have made significant progress \cite{qi2024computer,yu2021searching,9007047,wang2025foundation}. However, most of these methods are limited to close-set scenarios and lack the ability to handle unseen or novel gestures. In addition, they often rely on a large amount of annotated data, which is often difficult to obtain in practical applications.

Class-incremental gesture recognition is an important problem within the field of open-set recognition, where the model needs to continually learn and adapt to new gestures. However, catastrophic forgetting poses a significant challenge in this context. One straightforward approach to address this issue is to retain samples from old classes and mix them with samples of new classes during the subsequent incremental training of the model~\cite{rebuffi2017icarl, chaudhry2019tiny}. However, in real-world applications, past data may not always be fully accessible due to privacy and legal constraints associated with user data \cite{zhu2022self}. Furthermore, the limited memory space of devices poses another challenge, as the number of stored samples increases significantly with the addition of incremental tasks.

% plot a figure to show data-free class-incremental gesture recognition
% 4) Data-free class-incremental gesture recognition, advantages, prototype, variational
\begin{figure}[t]
	\centering
	\includegraphics[width=0.46\textwidth]{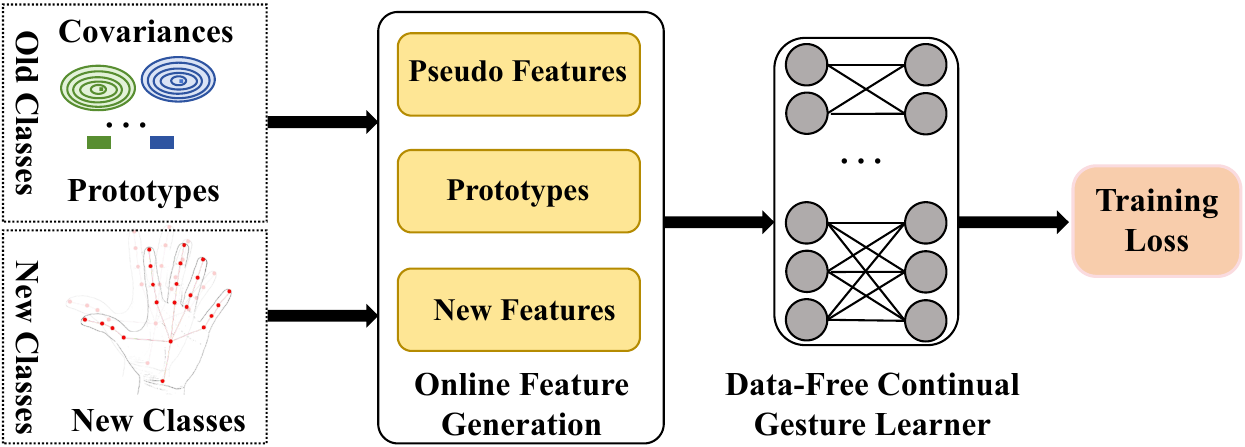}
	\caption{Illustration of the proposed data-free class-incremental gesture recognition. The main characteristics of this approach are generating pseudo features in an online fashion during the batch training of new class samples, leveraging prototypes and covariance matrices from old classes.}
	\label{fig:motivation}
	% \vspace{-10pt}
\end{figure}

Data-free class-incremental learning addresses the problems of data privacy concerns and limited memory space. Typical data-free approaches use model inversion to generate pseudo samples \cite{yin2020dreaming, aich2023data} or pseudo features \cite{petit2023fetril, liu2020generative} of old classes, which would eliminate the need to retain a large generator and only the previously trained network. Recently, Aich et al. \cite{aich2023data} introduce a data-free approach for class-incremental hand gesture recognition, leveraging model inversion to generate pseudo samples based on eigenvectors of the covariance matrix and the support vectors of the classifier. However, the samples obtained through model inversion for each task still need to be retained. Notably, this approach stores the inverted samples rather than the original samples, making it highly inefficient. Despite the significant computational cost of model inversion, the requirement for vast storage to retain these inversion samples escalates as the number of continual tasks increases.

% Data-free class-incremental learning refers to a learning paradigm where a model is trained to incrementally learn new classes without access to any previously encountered data samples. This approach addresses the problems of data privacy concerns and limited memory space. Typical data-free approaches use model inversion to generate pseudo samples \cite{yin2020dreaming, aich2023data} or pseudo features \cite{petit2023fetril, liu2020generative} of old classes, which would eliminate the need to retain a large generator and only the previously trained network. For example, DeepInversion \cite{yin2020dreaming} samples the initial sample from a normal distribution and generates a pseudo sample by making the distribution of its output in the model as similar as possible to past class labels. FeTrIL \cite{petit2023fetril} utilizes the geometric relationship between new and past class prototypes and new sample features to generate pseudo features. 

Our objective is to efficiently tackle class-incremental gesture recognition using a data-free approach, as shown in ~\ref{fig:motivation}. We focus on generating effective pseudo features for old classes, rather than pseudo samples, thereby significantly reducing computational costs \RED{during training}. To enhance the representation of past sample distributions, we introduce class prototypes and covariance matrices for old classes. During the batch training of new class samples for each task in continual learning, pseudo features, prototypes, and new features are concurrently fed into the classifier, working together to address the issue of catastrophic forgetting.
% We incorporate a consistency constraint between the weights of old classes in the classifier and their corresponding prototypes. This constraint, coupled with pseudo feature generation, effectively addresses the issue of catastrophic forgetting in a seamless manner.

To this end, we introduce a framework called Prototype-Guided Pseudo Feature Replay (PGPFR) tailored for class-incremental gesture recognition. The PGPFR comprises four crucial components: Pseudo Feature Generation with Batch Prototypes (PFGBP), Variational Prototype Replay for old classes, Truncated Cross-Entropy for new classes and Continual Classifier Re-Training. 
To generate a diverse set of pseudo features, the PFGBP combines class prototypes from old classes, features of new classes, and batch-specific class prototypes of new classes, which are computed within each batch of data, utilizing efficient geometric transformations. 
To strike a balance between overconfidence and uncertainty, temperature sharpening is used to control the steepness of probability predictions of pseudo features. The Variational Prototype Replay is built on the assumption that the classifier ought to precisely predict the class associated with the prototype of a particular old class. A plain prototype replay loss is introduced, followed by the Variational Prototype Replay that incorporates covariances to enhance robustness. The Truncated Cross-Entropy loss aims to mitigate the influence of domain differences of the classifier~\cite{smith2021always}. Lastly, we exploit the classifier re-training strategy for class-incremental gesture recognition.
% , ensuring the stability of the generated pseudo features and preventing overfitting to the new classes.
It should be noted that the spatial complexity of our model is low, since the pseudo features generated by our approach are discarded immediately after being used in each batch, eliminating the need for their retention.

In summary, the main contributions are listed as follows:
\begin{itemize} 
	\item To protect data privacy and improve efficiency, we introduce an efficient Prototype-Guided Pseudo Feature Replay framework for data-free class-incremental learning. 
	\item We propose a versatile module of Pseudo Feature Generation with Batch Prototypes that utilizes efficient operations and batch prototypes to generate pseudo features and pseudo labels for old classes in an online fashion.
	\item We propose Variational Prototype Replay which leverages class prototypes and covariance matrices to dynamically update the classifier weights of old classes. Additionally, we introduce Continual Classifier Re-Training strategy to further the problem mitigate catastrophic forgetting.
	% \textcolor{red}{Variational Prototype Replay  for Old Classes} which leverages class prototypes and covariance matrices to dynamically update the classifier weights of old classes while ensuring they remain aligned with their respective class prototypes.
	% \item We design a Truncated Cross-entropy for New Classes to mitigate the domain difference between real data and pseudo features. 
	% \item Our approach significantly outperforms existing state-of-the-art methods by 11.8\% and 12.8\% on the SHREC 2017 3D and the EgoGesture 3D benchmarks, respectively.
\end{itemize}

\section{Related work}
\subsection{Video-Based Hand Gesture Recognition}
Hand gestures primarily consist of various combinations of fingers. Earlier works employed shape matching to measure the dissimilarity among different hand shapes. Deep learning methods learn representations directly from raw data and perform recognition in an end-to-end manner. O Köpüklü et al. \cite{kopuklu2019real} introduce a novel hierarchical structure for neural networks, facilitating the efficient deployment of offline-trained CNN architectures using sliding window techniques. The attention mechanism can enable the model to focus on key areas related to gesture recognition in videos or images, thereby improving recognition performance. Consequently, Dhingra et al. \cite{dhingra2019res3atn} propose an end-to-end trainable 3D attention residual neural network, which stacks attention blocks and can adaptively change attention features. 
In order to better integrate multimodal information, Elboushaki et al. \cite{elboushaki2020multid} introduce an enhanced motion representation with multimodal fusion strategies at both feature and decision levels. Avola et al. \cite{avola20223d} employ an enhanced stacked hourglass network for multitasking, generating 2D heatmaps and hand contours, and predicting 3D hand joint positions and meshes. 

Recently, Li et al. \cite{li2023learning} propose an analytical framework based on the information bottleneck theory to reduce the interference of gesture irrelevant factors by optimizing feature coding. Although these methods achieve satisfactory performance on different gesture datasets, they rely on the RGB-D video as input. This dependency causes the model susceptible to factors like background noise, illumination variations, and viewpoint changes \cite{aich2023data}, ultimately leading to a degradation in model performance.

\subsection{3D Hand Gesture Recognition}
Similar to skeleton-based human action recognition~\cite{wang2025foundation,wang2018beyond,bian2021structural,hao2021hypergraph}, 3D hand gesture recognition refers to the understanding of hand gestures by capturing and analyzing the motion information of human joints. The 3D skeleton sequence is typically utilized as the input due to its inherent advantages in robustness and lightness. 
Nguyen et al. \cite{nguyen2019neural} introduce a novel neural network that learns the positive semi-definite matrix from skeletal data and constructs a novel layer using a variant of random gradient descent on the Stiefel manifold. De Smedt et al. \cite{de2019heterogeneous} argue that hand skeleton data can be used to extract valid hand kinematic descriptors in gesture sequences, statistically and temporally encoded via a Fisher kernel and a multilevel temporal pyramid. Liu et al. \cite{liu2020decoupled} argue that gestures can be decoupled into hand pose changes and hand movements and model them separately. Analogously, Guo et al. \cite{guo2021normalized} propose a new edge variation graph to divide each neighborhood of the central node into three groups of physical neighbors, temporal neighbors, and variational neighbors. 

Graph convolutional networks have proven to be effective in gesture recognition. 
Song et al. \cite{song2022dynamic} use a spatial and a temporal graph convolutional layer to learn the relationship between hand joints and to capture temporal features, respectively. To prevent spatial information and temporal information from influencing each other to produce fuzzy semantics, Liu et al. \cite{liu2023temporal} propose a framework to capture the spatiotemporal dependence between the skeleton joints by calculating the adjacency matrix of channel dependence and time dependence corresponding to different channels and frames. Transformer also plays an important role in extracting correlations between different nodes and different frames. Zhao et al. \cite{zhao2023spatial} utilize a Spatial-Temporal Synchronous Transformer to make the model focus on extracting spatial-temporal correlations simultaneously. 
Distinct from prior works, we focus on the challenging setting of class-incremental gesture recognition.

\subsection{Data-Free Class-Incremental Learning}
\RED{Similar to incremental clustered lifelong learning~\cite{sun2020representative}, generalized lifelong spectral clustering~\cite{sun2021and} and continual image generation~\cite{sun2024create,liu2025museummaker}, data-free class-incremental learning aims to mitigate catastrophic forgetting without relying on full access to previous data.}
Replay-based incremental learning methods \cite{rebuffi2017icarl, chaudhry2019tiny} face issues of user data privacy and linear increase in sample storage space with the increasing number of incremental tasks. Knowledge distillation methods \cite{li2017learning, dhar2019learning, rannen2017encoder} use a typical student-teacher architecture to distill the knowledge of past tasks from the teacher model into the student model. 
To transfer past knowledge to the model under new tasks without retaining past samples or models, data-free approaches \cite{shin2017continual, robins1995catastrophic, cong2020gan} utilize the Generative Adversarial Network (GAN) model structure to generate pseudo samples to mitigate catastrophic forgetting. The GAN model needs to be saved throughout the entire class-incremental learning process and has much larger computation and storage space compared to the classification model. 

Recent works use model inversion to generate past samples or features that are mixed with new class samples for model training, such as DeepInversion \cite{yin2020dreaming}, ABD \cite{smith2021always}, and BOAT-MI \cite{aich2023data}. Specifically, ABD \cite{smith2021always} improves DeepInversion \cite{yin2020dreaming} by proposing local cross entropy for new class samples and importance-weighted feature distillation for features. BOAT-MI \cite{aich2023data} stores the distribution information of past class samples and proposes a combination of SV-based and Proto-based methods to explore pseudo features of old classes in deep feature space. Nevertheless, the computational cost of BOAT-MI is considerable. 
Another disadvantage of BOAT-MI is the need to retain the pseudo samples obtained by model inversion for each task for subsequent incremental tasks, resulting in a sharp increase in the storage space of inversion samples with the increase of tasks. Our focus is data-free class-incremental gesture recognition without retaining pseudo data, and the objective is to design an efficient pseudo-feature-based approach that fully leverages the advantages of prototypes.

\begin{figure*}
	\centering
	\includegraphics[width=1.0\textwidth]{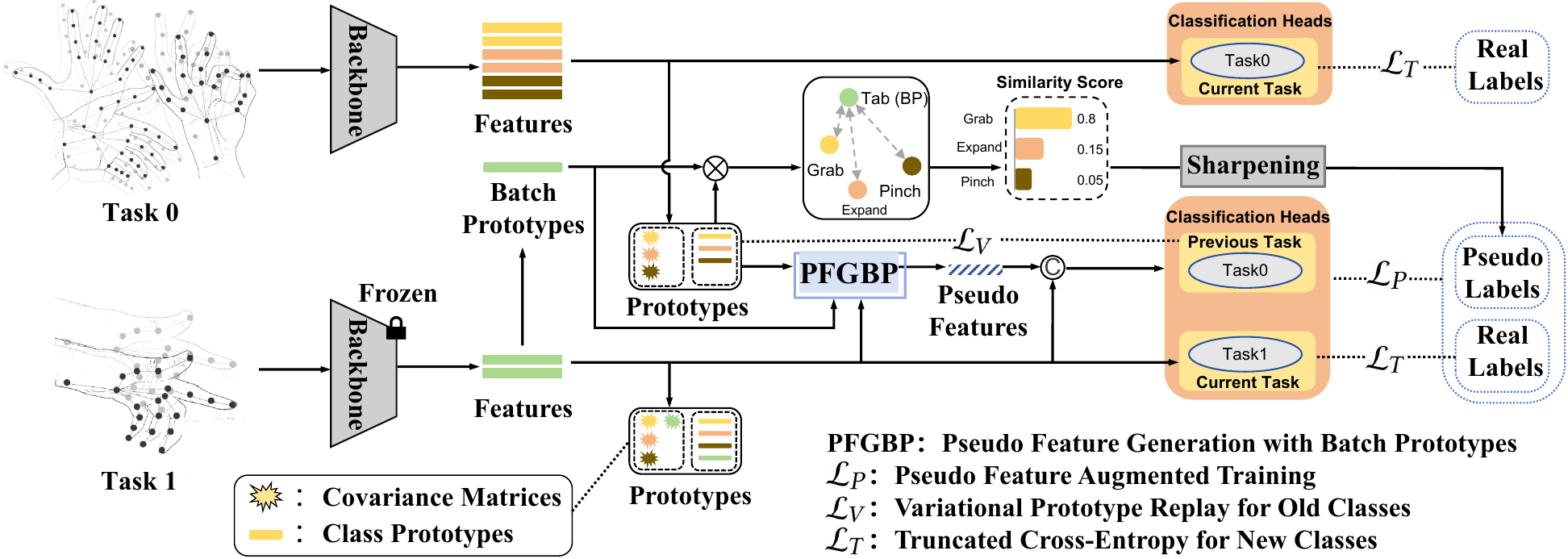}
	\caption{Framework of the proposed data-free class-incremental gesture recognition with
		Prototype-Guided Pseudo Feature Replay (PGPFR). Backbone is fixed after being trained on $\mathcal{T}_0$. In each subsequent class increment task $\mathcal{T}_i$, we use class prototypes and covariance matrices to generate pseudo features of old classes through PFGBP, only adjusting classification heads. For the acquisition of new class knowledge, we minimize  $\mathcal{L}_T$ to learn the decision boundaries of the new class; For preserving knowledge of old classes, we minimize $\mathcal{L}_P$ and $\mathcal{L}_V$ to strengthen the decision boundaries of old classes.} 
	\label{fig:method}
\end{figure*}

\section{Methodology}
For Class-Incremental Learning, the model continuously learns $N$ incremental tasks 
\{$\mathcal{T}_0,\mathcal{T}_1,...,\mathcal{T}_i,...,\mathcal{T}_N $\}. The first task  
$ \mathcal{T}_0 $ requires $k$ classes, and subsequently, for each incremental task , $d$ new classes $ \mathcal{C}_n $ are added to the existing classes $ \mathcal{C}_o $ from all previous old tasks. During each incremental task $ \mathcal{T}_i $, a specific dataset $ \mathcal{D}_i $ for $ \mathcal{T}_i $ is provided for model training. 
%, and it is necessary to ensure that the datasets for each different task period do not intersect
In the context of data-free class-incremental learning, the model is invisible to the training set that has already been seen in the past and can only be trained on $ \mathcal{D}_i $  for $ \mathcal{T}_i $. The critical issue with data-free class-incremental learning is catastrophic forgetting, a phenomenon where the decision boundary of the classifier gradually becomes blurred as more incremental tasks are introduced, ultimately resulting in a substantial decline in model performance.

We propose an effective Prototype-Guided Pseudo Feature Replay (PGPFR) framework. We strengthen the decision boundaries of old classes by introducing Pseudo Feature Generation with Batch Prototypes (PFGBP) and Variational Prototype Replay for old classes. In addition, we employ Truncated Cross-Entropy to learn decision boundaries for new categories, alongside the Continual Classifier Re-Training strategy, which addresses the challenges of overfitting to instances of new classes and underfitting to those of old classes. 
% On the other hand, we not only utilize General cross entropy but also utilize Truncated cross entropy to learn decision boundaries for new categories. It should be noted that the backbone is fixed after it is trained on $\mathcal{T}_0$, which means that it no longer changes with the variation of the CIL task and only classification is re-trained in the sequent task. 

For convenience, we divide the model into two parts: a backbone network $ f(\cdot)$ and a classifier $ \varphi(\cdot) $. DG-STA \cite{chen2019construct} is employed as the backbone to effectively extract 3D gesture features, and a fully connected layer serves as the linear classifier for gesture recognition. The overall architecture of the proposed method is illustrated in Figure~\ref{fig:method}.

\subsection{Pseudo Feature Generation with Batch Prototypes}\label{sec:batch_prototype}
To strengthen the decision boundary of the old classes, Pseudo Feature Generation with Batch Prototypes (PFGBP) is designed to generate pseudo features of old classes. Inspired by~\cite{petit2023fetril}, we obtain pseudo features by performing simple geometric translations of prototypes of old classes in the feature space. The pseudo feature generated by the PFGBP is represented as:
\begin{equation}
	\hat{f}(x_p) = f(x_n) + \mu_p - \hat{\mu}_n, \label{con:eq1}
\end{equation}
where $n$ and $p$ are indexes of new class and previous classes, respectively, $x_n$ is a sample of the new class, $\mu_p$ is the prototype of the $p$-th class, $\hat{\mu}_n$ is the estimated prototype for the $n$-th class in a batch of data, $\hat{f}(x_p)$ is the generated pseudo feature corresponding to the $p$-th class.

The inclusion of all new class samples in the input guarantees that pseudo features and new class samples exhibit a comparable distribution, thereby addressing the issue of data imbalance. \RED{Since the pseudo features are generated together with all new class samples, they naturally align with the same underlying feature distribution, thereby balancing the contributions of old and new classes during training.
Simple geometric translations cause the pseudo features to reside in the feature space close to the prototype of the past class, indicating the validity of pseudo features. }

The pseudo labels of old classes corresponding to the generated pseudo features are selected as follows. For each batch of new class samples, we first calculate the batch class prototype, denoted as $\hat{\mu}_n$. Then, the similarities between $\hat{\mu}_n$ and prototypes of old classes are computed. The most similar past class prototype is chosen as $\mu_p$, and the corresponding class serves as the pseudo label $\hat{y}$. The formula for assigning labels to pseudo features can be expressed as:
\begin{equation}
	\hat{y} = \mathrm{argmax}_{0 \leq p \leq \mathcal{N}_o} \mathrm{sim}(\hat\mu_n, \mu_p),
	\label{con:eq2}
\end{equation}
where $\mathcal{N}_o$ is a number of old classes, and $\mathrm{sim}(\cdot) $ denotes the cosine similarity function.

It should be noted that we utilize batch class prototypes instead of the class prototypes derived from all samples within a given class. By leveraging batch class prototypes, we can generate a more diverse set of pseudo features. Consider the scenario that $\mathcal{T}_n$ comprises only a single new class; in such a case, the pseudo label $\hat{y}$ and its corresponding $\mu_p$ would remain fixed to the same old category throughout the $n$-th task. This constraint limits the diversity of pseudo features and ultimately prevents the model from effectively revisiting knowledge from other older classes. Compared with the original prototype, the batch class prototype could better represent the characteristics of the individual data in the batch. Namely, using all samples to compute the prototype may capture some noise or unnecessary details within the class, while computing the prototype using only a batch can reduce this risk, thereby generating a more representative prototype. The specific generation process of pseudo features and pseudo labels is outlined in Algorithm \ref{alg:algorithm}.

The pseudo features $\hat{f}(x_p)$ are concatenated with the features of the new class samples $f(x_n)$ to form $ \tilde{Z} = [\hat{f}(x_p), f(x_n)]$. The corresponding labels $\hat{y} $ and y are also concatenated to form $ \tilde{Y} = [\hat{y}, y]$. Finally, $\tilde{Z}$ and $ \tilde{Y}$ are fed into the classifier, and the cross entropy between the predicted distributions and the labels are:
\begin{equation}
	\mathcal{L}_P = \sum_{z_k\in \tilde{Z}, y_k \in \tilde{Y}} l(z_k, y_k) \label{con:eq3},
\end{equation}
where $l(z,y)$ is defined as
\begin{equation}
	l(z, y) =  \begin{cases}
		-\sum_{i=1}^{\mathcal{N}_o} y_i \log (\mathrm{softmax}(\varphi(z) / R))   \text{, if } z \in \hat{f}(x_p) \\
		-\sum_{i=1}^{\mathcal{N}_o} y_i \log (\mathrm{softmax}(\varphi(z)))     \text{,  else } z \in f(x_n),
	\end{cases}   \label{con:eq4}
\end{equation}
where $R$ is a temperature parameter to sharpen the predictive distribution of pseudo features, y is a one-hot vector corresponding to a real label or pseudo label and $\mathrm{softmax}(\cdot)$ is an activation function that normalizes classified logits. 

It should be noted that temperature sharpening adjusts the probability distribution output by the softmax function through the parameter \( R \). When \( R < 1 \), the distribution becomes sharper, meaning the probabilities for the most likely classes increase while the probabilities for less likely classes decrease. This strategy could reduce the model's sensitivity to uncertain classes, mitigating the negative impact of noise on training. 

% As noise is inevitable during the generation of pseudo labels, 
% During the generation of pseudo-labels, noise inevitably leads to less accurate labels. Temperature Sharpening adjusts the probability distribution output by the softmax function through a temperature parameter \( R \). When \( R < 1 \), the distribution becomes sharper, meaning the probabilities for the most likely classes increase while the probabilities for less likely classes decrease. By modifying the probability distribution, the model becomes more confident in the most probable class, resulting in a sharper distribution. Temperature Sharpening effectively reduces the model's sensitivity to uncertain classes, mitigating the negative impact of noise on training. It tends to strengthen the prediction for a particular class, making the pseudo-labels more reliable. As Temperature Sharpening makes class predictions more definite, the model focuses more on high-confidence predictions during subsequent training, reducing the accumulation of errors caused by inaccurate pseudo-labels. This helps to improve the model's generalization ability, especially in scenarios involving incremental learning or unsupervised learning

\begin{algorithm}  
	\caption{The generation process of pseudo features and pseudo labels}
	\label{alg:algorithm}
	\textcolor{lightgray}{\# O: the number of past class prototypes}\\
	\textcolor{lightgray}{\# B: represents the number of batches}\\
	\textcolor{lightgray}{\# T: Transposition operation}\\
	\textbf{Input}: a batch of samples $x_n = {x}_{1:B}$, the feature extraction network $f(\cdot)$ in the current task, prototype set $\mathcal{P}_{\text{old}} = {\{\mu_i\}_1^O}$.\\
	\textbf{Output}: a set of pseudo features $\mathcal{F}_{\text{pseudo}}$ and a set of pseudo labels $\mathcal{L}_{\text{pseudo}}$.
	
	\begin{algorithmic}[1]
		\STATE \textcolor{lightgray}{\# Initialize the set of pseudo features and pseudo labels}
		\STATE $\mathcal{F}_{\text{pseudo}} \gets \{\}, \mathcal{L}_{\text{pseudo}} \gets \{\}$
		\STATE \textcolor{lightgray}{\# Utilize features the extraction network $f(\cdot)$ extract features $f(x_n)$ from $x_n$ }
		\STATE $x_n \overset{f}{\to} f(x_n)$
		\STATE \textcolor{lightgray}{\# Calculate the means of $f(x_n)$ corresponding to labels to obtain Batch Class Prototypes $\hat\mu_n$}
		\STATE $f(x_n) \overset{Mean}{\to} \hat\mu_n$
		\STATE \textcolor{lightgray}{\# Calculate the similarity matrix $C$ between $\hat\mu_n$ and $\mathcal{P}_{\text{old}}$}
		\STATE $ C \gets \hat\mu_n \cdot \mathcal{P}_{\text{old}}^T$
		\STATE \textcolor{lightgray}{\# Select the old category p that is most similar to $\hat\mu_n$}
		\STATE $p \gets \arg\max_j \, C_{ij}$
		\STATE \textcolor{lightgray}{\# Generate pseudo features using Eq.(1)}
		\STATE $\hat{f}(x_p)  \gets  f(x_n) + \mu_{p} - \hat{\mu}_n$ 
		\STATE \textcolor{lightgray}{\# Merge $\hat{f}(x_p)$ and $p$ into $\mathcal{F}_{\text{pseudo}}$ and $\mathcal{L}_{\text{pseudo}}$}
		\STATE $\mathcal{F}_{\text{pseudo}} \gets \mathcal{F}_{\text{pseudo}} \cup \{\hat{f}(x_p)\}$
		\STATE $\mathcal{L}_{\text{pseudo}} \gets \mathcal{L}_{\text{pseudo}} \cup \{p\}$
	\end{algorithmic}
\end{algorithm}  

\subsection{Variational Prototype Replay  for Old Classes}
%In the process of continuously learning new class knowledge, the decision boundaries of old classes will suffer catastrophic damage. This requires us to regularize the classifier to a certain extent. If we can access samples of old classes, then this will not be a problem. However, this contradicts the focus of this article. Although the samples of old classes are not visible, we can obtain their distribution information (class mean and covariance). Therefore, we decided to use the distribution information of samples to correct the classifier and alleviate the damage to the decision boundaries of old classes.
% We feed the class prototype as information of old classes to the classifier to strengthen the decision boundary, which can enable the classifier to better distinguish the differences between different categories. 
For old classes, the prototype of a particular class should be as close as possible to the classifier's weight for that class, while being as far away as possible from the classifier's weights for other classes. In other words, the classifier should accurately predict the class for the prototype of a given old class. We define prototype replay loss to enable the classifier to better distinguish the differences between different prototypes of old classes. The loss is formulated as:
\begin{equation}   
	\mathcal{L}_{Proto} = \frac{1}{\mathcal{N}_o}\sum^{\mathcal{N}_o}_{k = 1}-\log(\frac{e^{w^{T}_{k}\mu_{k}+b_k}}{\sum^{\mathcal{N}_o}_{c = 1}e^{w^T_c\mu_k+b_c}}),\label{con:eq5}
\end{equation}
where $\mathcal{N}_o$ is a number of old classes, $w_k$ is the classification weight for the  $k$-th class in the fully connected layer, $b_k$ is the bias vector, and $\mu_k$ is the prototype of the $k$-th class.

The original prototype lacks a measure of sample distributions of the given class. Samples closer to class prototypes in the feature space are more likely to be classified correctly, while those farther away may be misclassified, which leads to a single solution of the intra-class features. We employ the class prototypes and covariance matrices, which characterizes each class as a distribution rather than a singular point in the latent space. To enhance the robustness and generalization, we incorporate covariances and penalize the predictions of class prototypes with a regularization term in the denominator. The Variational Prototype Replay loss is formulated as follows:
\begin{equation}   
	\mathcal{L}_{V} = \frac{1}{\mathcal{N}_o}\sum^{\mathcal{N}_o}_{k = 1}-\log(\frac{e^{w^{T}_{k}\mu_{k}+b_k}}{\sum^{\mathcal{N}_o}_{c = 1}e^{w^T_c\mu_k+b_c+\gamma (w_c-w_k)^T C_k(w_c-w_k)}}), \label{con:eq6}
\end{equation}
where $\mathcal{N}_o$ is the number of old classes, $\mu_{k}$ is the class prototype of the $k$-th class, $C_k$ is the covariance matrix of samples of the $k$-th class, and $\gamma$ is a hyperparameter. 
% we introduce a regularization term by incorporating covariance information. 
% $\gamma(w_c-w_k)^T\sum_k(w_c-w_k)$, 
% 由于协方差矩阵一定是一个对称矩阵，那么正则项是大于或等于0的。从最小化损失值的角度来看，我们实际上是增大了公式的分母部分，要想使得损失值更小，那么必须增大分子部分，这使得正确类别的预测概率在logits中的占比更大。然而，分子部分正是我们需要分类器去分类正确类别的概率，这会加强分类器对正确类别的预测。
% always greater than or equal to 0. From the perspective of minimizing the loss value, we effectively increase the denominator of \myref{con:eq4}. To minimize the loss value further, we must increase the numerator, which amplifies the proportion of the correct class's predicted probability in the output. The numerator represents the probability of the classifier correctly classifying the class prototype, which strengthens the classifier's prediction for the correct class. From this perspective, our penalty term is also reasonable. 

As the covariance matrix is symmetric, the penalty term in the denominator is positive. To minimize the loss $\mathcal{L}_{V}$, the optimizer would increase the numerator which represents the likelihood of the classifier accurately classifying the class prototypes, thereby improving the consistency between the classifier and the prototypes.

\subsection{Truncated Cross-Entropy for New Classes}
As mentioned in~\cite{smith2021always}, there exists a domain difference between real data and pseudo features generated by model inversion. When pseudo features and real data are mixed for model training, the model tends to classify the mixed samples as either real or synthetic, rather than discriminating them based on the semantic information encoded within the features. To mitigate the influence of domain differences of the classifier, we introduce an additional classifier solely dedicated to the new classes. This classifier, termed Truncated Cross-entropy for New Classes, is formulated as follows:
% |y \in \mathcal{T}_n
\begin{equation}
	\mathcal{L}_T = \sum_{x\in \mathcal{D}^n} \sum_{y \in \mathcal{T}_n}\ - y\mathrm{log}(\mathrm{softmax}(\varphi^n(f(x))))), \label{con:eq7}
\end{equation}
where $\varphi^n(\cdot)$ is the classification header of the $n$-th task, $f(\cdot)$ is the feature extractor.

\subsection{Continual Classifier Re-Training}
The overall training objective loss is:
\begin{equation}
	\mathcal{L} = \mathcal{L}_{P} + \mathcal{L}_{V} + \mathcal{L}_{T}.  \label{con:eq8}
\end{equation}
% addresses the challenges of overfitting to instances of new classes and underfitting to those of old classes. 

During training, we design Continual Classifier Re-Training strategy, which is inspired by classifier re-training~\cite{kang2019decoupling} in long-tail learning. However, in class incremental learning, this training strategy is rarely studied. Specifically, we fix the backbone after training it on the first task and keep the backbone parameters unchanged throughout subsequent incremental tasks. For each subsequent incremental task, we only re-train the classifier to adjust the decision boundary. This strategy guarantees the stability of extracted features, preserving a consistent feature space for pseudo feature generation. Fixing the backbone allows for the acquisition of a more generalized feature representation, which aids the classifier in adapting to new tasks. It also avoids overfitting to new classes and retains feature information from old classes, helping to mitigate catastrophic forgetting. 
\RED{We choose classifier re-training because it neither introduces additional parameters nor requires extra training data.}

\section{Experiment}
\subsection{Dataset}

The \textbf{SHREC 2017 3D} dataset, derived from Online-DHG, includes 14 gesture sequences performed by 28 participants, totaling 2800 sequences. It captures both coarse and fine gestures using 22 hand joints in 2D and 3D space, Each of which is represented by a single coordinate. The skeletal structure is represented by the coordinates of 22 joints in both 2D depth image space and 3D world space, constituting a complete hand representation. These 22 joint points include 1 palm, 1 wrist, and 4 joints per finger, totaling 5 fingers.

The \textbf{EgoGesture 3D} dataset is a first-person perspective dataset with 24161 gesture samples recorded by 50 subjects across various environments, extracted from the original dataset EgoGesture through the MediaPipe \cite{zhang2020mediapipe} API. This dataset encompasses 83 types of static or dynamic gestures, primarily focusing on interactions with wearable devices.
It complements the SHREC 2017 dataset, offering a broader environmental, accounting for lighting and background changes, as well as gestures performed with one hand and both hands, for validating gesture recognition methods. 
% We segmented these two datasets to demonstrate the effectiveness of our method in the experimental setting of Data-Free class incremental learning according to \cite{aich2023data}. 

\begin{table*}[t]
	\caption{Comparison with state-of-the-art approaches of class-incremental gesture recognition on the EgoGesture 3D dataset.}
	\resizebox{\linewidth}{!}{
		\begin{tabular}{l|c|cc|cc|cc|cc|cc|cc|cc}
			\toprule
			\multirow{2}{*}{Method}           & Task 0                 & \multicolumn{2}{c|}{Task~1}    & \multicolumn{2}{c|}{Task~2}    & \multicolumn{2}{c|}{Task~3}    & \multicolumn{2}{c|}{Task~4}    & \multicolumn{2}{c|}{Task~5}    & \multicolumn{2}{c|}{Task~6}  & \multicolumn{2}{c}{Mean}   \\     \cmidrule{2-16} 
			
			& G↑(\%)                    & G↑(\%)            & IFM↓         & G↑(\%)            & IFM↓         & G↑(\%)            & IFM↓         & G↑(\%)            & IFM↓         & G↑(\%)            & IFM↓         & G↑(\%)            & IFM↓      & G↑(\%)       & IFM↓    \\ 
			\midrule	
			Oracle             &75.8 & 75.8 & -- &75.8& -- &75.8& -- &75.8& -- &75.8& -- &75.8 & --  & 75.8 & -- \\ 
			\midrule				   
			Base \cite{li2017learning}               & 78.1 & 60.4          & 13.5         & 18.9          & 63.3         & 9.3           & 82.4         & 8.0           & 84.3         & 5.9           & 88.5         & 5.9           & 88.3  &18.1  & 70.0         \\
			Fine-tuning \cite{li2017learning}       &78.1                       & 59.2          & 10.4         & 15.9          & 66.5         & 9.3           & 82.2         & 7.7           & 84.4         & 5.2           & 89.7         & 5.0           & 90.1  &17.1 &70.6        \\
			Feature extraction \cite{li2017learning} &78.1                      & 69.8          & 14.5         & 60.8          & 17.2         & 51.5          & 30.4         & 46.4          & 33.7         & 41.2          & 39.4         & 36.8          & 42.9  &51.1 &29.7        \\
			LwF \cite{li2017learning}               &78.1                       & 69.1          & 0.1          & 43.0          & 27.3         & 18.8          & 67.3         & 11.3          & 78.7         & 6.5           & 87.5         & 6.3           & 87.7  &25.8 &58.1       \\
			LwF.MC \cite{smith2021always}            &78.1                       & 36.8          & 9.8          & 24.2          & 41.3         & 17.0          & 64.6         & 12.8          & 73.3         & 10.2          & 78.3         & 9.7           & 80.2  &18.4 &57.9        \\
			DeepInversion \cite{yin2020dreaming}     & 78.1                      & 68.1          & 14.1         & 44.3          & 32.2         & 24.7          & 59.2         & 16.2          & 71.2         & 11.6          & 78.8         & 10.0          & 81.0        &29.1 &56.1 \\
			ABD \cite{smith2021always}               &78.1                       & 68.8          & 14.8         & 61.1          & 17.1         & 54.3          & 26.2         & 49.0          & 31.1         & 43.2          & 37.5         & 39.0          & 40.6   &52.6 &27.9      \\
			R-DFCIL \cite{gao2022r}           & 78.1                      & 70.3          & 3.0          & 61.4          & 4.1          & 53.2          & 14.9         & 46.1          & 17.8         & 39.2          & 35.0         & 35.2          & 36.3        &50.9 &18.5 \\
			BOAT-MI \cite{aich2023data}           &  78.1                     & 75.3          & 4.6 & 70.9          & \textbf{7.3} & 64.0          & 15.1         & 58.6          & 16.3         & 52.3          & 26.8         & 45.8          & 32.2       &61.2 &17.7  \\
			\midrule	
			PGPFR (ours)     & \textbf{79.3}         & \textbf{76.8} & \textbf{4.3}          & \textbf{73.8} & 11.9          & \textbf{72.2} & \textbf{3.7} & \textbf{71.0} & \textbf{1.8} & \textbf{69.8} & \textbf{0.6} & \textbf{68.4} & \textbf{2.1}  & \textbf{73.0} & \textbf{4.1}\\  
			\bottomrule
		\end{tabular}
	}
	\label{tab:table1}
\end{table*}

\begin{table*}[t]
	\caption{Comparison with state-of-the-art approaches of class-incremental gesture recognition on the SHREC 2017 3D dataset.}
	\resizebox{\linewidth}{!}{
		\begin{tabular}{l|c|cc|cc|cc|cc|cc|cc|cc}
			\toprule
			\multirow{2}{*}{Method}           & Task 0                 & \multicolumn{2}{c|}{Task~1}    & \multicolumn{2}{c|}{Task~2}    & \multicolumn{2}{c|}{Task~3}    & \multicolumn{2}{c|}{Task~4}    & \multicolumn{2}{c|}{Task~5}    & \multicolumn{2}{c|}{Task~6}  & \multicolumn{2}{c}{Mean}   \\     \cmidrule{2-16} 
			
			& G↑(\%)                    & G↑(\%)            & IFM↓         & G↑(\%)            & IFM↓         & G↑(\%)            & IFM↓         & G↑(\%)            & IFM↓         & G↑(\%)            & IFM↓         & G↑(\%)            & IFM↓      & G↑(\%)       & IFM↓    \\ 
			\midrule
			Oracle          & 89.4& 89.4&  -- &89.4& --  &89.4& -- &89.4& -- &89.4& -- &89.4& --  & 89.4 & --  \\
			
			\midrule
			Base \cite{li2017learning}              &90.5 & 78.3          & 7.8          & 53.3          & 29.6         & 30.0          & 53.7         & 27.9          & 56.0          & 12.2          & 78.3         & 11.9          & 78.6  &35.6 &50.7        \\
			Fine tuning \cite{li2017learning}       &90.5                       & 78.6          & 6.0          & 57.4          & 26.2         & 34.2          & 48.8         & 34.0          & 48.6          & 16.4          & 71.8         & 16.1          & 71.9  &39.5 &45.5       \\
			Feature extraction \cite{li2017learning} & 90.5                      & 79.9          & 9.3          & 69.3          & 12.6         & 62.3          & 19.1         & 56.5          & 24.1          & 50.6          & 24.3         & 45.1          & 32.4  &60.6 &20.3       \\
			LwF \cite{li2017learning}               &90.5                       & 79.8          & 9.3          & 64.1          & 20.2         & 31.9          & 50.2         & 29.1          & 53.1          & 13.1          & 76.8         & 11.5          & 79.3   &38.2 &48.2      \\
			LwF.MC \cite{smith2021always}              & 90.5                      & 56.0          & 10.9         & 32.6          & 39.5         & 21.8          & 58.5         & 19.3          & 61.6          & 16.7          & 53.3         & 16.1          & 45.4   &27.1 &44.9      \\
			DeepInversion \cite{yin2020dreaming}     &90.5                       & 79.9          & 4.3          & 65.9          & 14.9         & 53.1          & 29.5         & 49.5          & 32.8          & 34.2          & 47.7         & 32.1          & 49.7   &52.5 &29.8      \\
			ABD \cite{smith2021always}                &90.5                       & 78.8          & 4.0          & 64.6          & 12.6         & 54.3          & 20.3         & 53.2          & 24.6          & 46.1          & 20.0         & 40.4          & 23.3     &56.2 &17.5    \\
			R-DFCIL \cite{gao2022r}           &90.5                       & 78.7          & 3.3          & 65.5          & 4.5          & 54.4          & 20.5         & 49.8          & 26.9          & 41.5          & 25.0         & 38.6          & 33.3    &54.8 &18.9     \\
			BOAT-MI \cite{aich2023data}           &90.5                       & 83.7          & 4.5          & 76.0          & 7.3          & 71.4          & 7.0          & 69.4          & \textbf{13.3} & 64.1          & 9.5          & 58.1          & 11.2    &70.5 &8.8     \\
			\midrule
			PGPFR (ours)   & \textbf{91.2}         & \textbf{90.2} & \textbf{0.7} & \textbf{89.4} & \textbf{1.0} & \textbf{83.7} & \textbf{5.8} & \textbf{78.3} & 14.3         & \textbf{74.5} & \textbf{7.7} & \textbf{75.8} & \textbf{11.1}  & \textbf{83.3} 
			&\textbf{6.8} \\ 
			\bottomrule
		\end{tabular}
	}
	\label{tab:table2}
\end{table*}

\subsection{Experimental Setup}  \label{subsec:Experimental Setup}
The optimization strategy is Adaptive Moment Estimation (Adam) with a learning rate of 0.001. The batch size is 32. The initial training task covers 150 epochs, whereas each subsequent incremental training task comprises 100 epochs. An experiment is set up with a total of 7 incremental tasks. The temperature parameter $R$ of sharpening is set to 0.3, which can be adjusted with the variation of the dataset. The hyperparameter of the penalized term $\gamma$ is set to 1. The implementation is based on the PyTorch, and all experiments are performed on an NVIDIA RTX 4090 GPU with CUDA 12.0. The parameter count of our model is approximately 0.27M, which is light-weighted.

To assess the method, we utilize two types of test sets: one encompassing all visible categories and another consisting of current categories. For the SHREC 2017 3D dataset, \RED{we follow the 14-gesture setting.}
After the model is trained on the initial incremental task (comprising 8 categories), we keep the backbone constant, while 1 new category is added to the visible classes during each subsequent incremental task. Similarly, for the EgoGesture 3D dataset, the backbone remains fixed once the model is trained on the first incremental class task (containing 59 categories), with 4 new categories being added during each subsequent incremental class task.

% \subsection{Evaluation Metrics}
A critical issue in the development of incremental/continual learning systems lies in the disparity between the retention of information about previously learned tasks/classes and that of the newly introduced ones. As a result, a significant imbalance arises between the accuracy evaluated for current classes and the accuracy evaluated for all visible classes. We use the Instantaneous Forgetting Measure: 
\begin{equation}
	IFM = \frac{|L-G|}{L+G}\times 100,
\end{equation} proposed in \cite{aich2023data} to measure the balance between our model's retention of knowledge of past categories and its learning of knowledge of new categories. In particular, L represents the testing accuracy for new classes within the current task, while G (Global Accuracy) represents the testing accuracy for all visible classes up to the current task. The smaller the gap between L and G, the more balanced the model's ability to retain knowledge of old classes and acquire knowledge of new classes. In the extreme case where the accuracy of either L or G is 0, the value of IFM reaches its maximum, indicating that the model's capacity to balance these two abilities is severely limited. In the experimental tables, we primarily reported the evaluation indicators G (where a higher value is preferable) and IFM (where a lower value is desirable).

\begin{table*}[t]
	\caption{Ablation studies of the proposed method on the SHREC 2017 3D dataset. Our baseline is the ABD~\cite{smith2021always}. CCRT denotes continual classifier re-training. Sharpening and Batch Proto denote temperature sharpening of predicted logits of pseudo features and batch class prototypes in the PFGBP. For simplicity, PGPFR w/o $\mathcal{L}_{V}$ is shortened as w/o $\mathcal{L}_{V}$, etc.
	}
	\resizebox{\linewidth}{!}{
		\begin{tabular}{l|ccccccccccccccc}
			\toprule
			\multirow{2}{*}{Method} & Task 0 & \multicolumn{2}{c}{Task 1}    & \multicolumn{2}{c}{Task 2}    & \multicolumn{2}{c}{Task 3}    & \multicolumn{2}{c}{Task 4}    & \multicolumn{2}{c}{Task 5}    & \multicolumn{2}{c}{Task 6}     & \multicolumn{2}{c}{Mean}     \\ \cmidrule{2-16}
			& G↑(\%)    & G↑(\%)            & IFM↓         & G↑(\%)            & IFM↓         & G↑(\%)            & IFM↓         & G↑(\%)            & IFM↓         & G↑(\%)            & IFM↓         & G↑(\%)            & IFM↓          & G↑(\%)            & IFM↓         \\ 
			\midrule
			PGPFR (ours) & 91.2 & 90.2         &\textbf{0.7}&\textbf{89.4}&1.0&\textbf{83.7}&\textbf{5.8} &\textbf{78.3} &14.3         &\textbf{74.5} &\textbf{7.7} &\textbf{75.8} &\textbf{11.1}&\textbf{83.3} & 6.8  \\
			\hline
			w/o PFGBP & 91.2  & 89.0        & 2.9         & 86.8        & 4.0         & 80.6        & 3.9         & 76.5        & 7.7         & 73.7        & \textbf{5.9}         & 75.0        & 12.5        &81.8&\textbf{6.2} \\
			w/o Sharpening  &91.2 &90.1 &0.6&89.1&1.2&83.7&5.8&77.9&34.5&74.1&7.5&75.4&6.6&83.1& 9.4 \\
			w/o Batch Proto &91.2 &86.9 &2.1 &86.3 &5.8 &81.5 &\textbf{0.5} &77.0 &\textbf{3.5} &74.2 &\textbf{3.7} &72.9 &14.7 &81.4 &\textbf{5.0}\\ 
			\midrule
			w/o $\mathcal{L}_{V}$ & 91.2  & \textbf{90.6}       & 1.0         & 85.5        & 3.2         & 75.8        & \textbf{3.5}         & 70.4        & 7.7         & 66.8        & 6.2         & 60.1        & 20.6       &77.2&7.0 \\
			w/o $\mathcal{L}_{V}$ w/ $\mathcal{L}_{Proto}$  & 91.2  & \textbf{90.4} & 1.1          & 87.4          & \textbf{0.5} & 71.9          & 11.6         & 67.5          & \textbf{0.2} & 66.3          & 10.7         & 65.4          & 13.6          & 77.2          & \textbf{6.3} \\
			\midrule
			w/o $\mathcal{L}_{T}$ &91.2&90.1&\textbf{0.3}&89.2&\textbf{0.4}&82.8&10.8&77.6&11.1&\textbf{74.8}&10.7&75.5&\textbf{8.5}&83.0&7.0 \\
			\midrule
			Baseline & 90.5  & 78.8        & 4.0         & 64.6        & 12.6        & 54.3        & 20.3        & 53.2        & 24.6        & 46.1        & 20.0        & 40.4        & 23.3        & 56.2&17.5\\
			Baseline w/ CCRT & 91.2  & 83.6        & 6.9         & 77.5        & 11.1        & 63.0        & 12.0        & 66.9        & 17.2        & 58.0        & 15.4        & 48.8        & 34.4        &69.9&16.2 \\
			\bottomrule
		\end{tabular}
	}
	\label{tab:ablation}
\end{table*}

\subsection{Results of Gesture Recognition}
Table \ref{tab:table1} shows the comparison of our approach with state-of-the-art approaches on the EgoGesture 3D dataset. Tasks 0 to 6 represent the sequential stages of continuous data-free class incremental learning, while \textit{Mean} denotes the average result across all seven tasks. It can be observed that our method achieves an average increase of 11.8\% in Global Accuracy when compared to BOAT-MI \cite{aich2023data}, which is a recent state-of-the-art approach for data-free class-incremental gesture recognition. Furthermore, in terms of IFM, our method demonstrates an average decrease of 13.6. Particularly, for task 6, our approach outperforms BOAT-MI \cite{aich2023data} by 31.1 in terms of IFM. 

The results on the SHREC 2017 3D dataset are shown in Table~\ref{tab:table2}. Similarly, our approach significantly outperforms state-of-the-art methods. For the Global Accuracy, our approach beats BOAT-MI \cite{aich2023data} by 12.8\%. 
% demonstrating its superior performance in continuous data-free class incremental learning.
The results clearly demonstrate the remarkable adaptability of our approach in accommodating new classes while preserving the decision boundaries of previously learned ones. 

% Table \ref{tab:table1} and Table \ref{tab:table2} present a comparative analysis of our approach with state-of-the-art methods. Tasks 0 to 6 represent the sequential stages of continuous data-free class incremental learning, while 'Mean' indicates the average result across all seven tasks. It can be observed that our method achieves an increase of 10.8\% and 12.8\% in terms of Global Accuracy on average for both EgoGesture 3D (Table \ref{tab:table1}) and SHREC 2017 3D (Table \ref{tab:table2}), respectively. Especially on task6 with the most severe forgetting, our method achieves an increase of 21.5\% and 17.7\%, respectively. This indicates that our method possesses a stronger ability to learn new class knowledge and retain decision boundaries for past old classes. From the perspective of IFM, our model shows a decrease of 13.9 and 2.0 on average, respectively. Especially on task6 with the most severe forgetting, our method presents a decline of 30.9 and 0.1, respectively. This indicates that our model possesses a stronger ability to balance the retention of past class knowledge and the learning of new class knowledge. In summary, our model significantly exhibits superior performance compared to the aforementioned state-of-the-art methods.

\begin{figure}[tbp]
	\centering
	\subfigure[Accuracy of new classes.]{
		\includegraphics[width=0.22\textwidth]{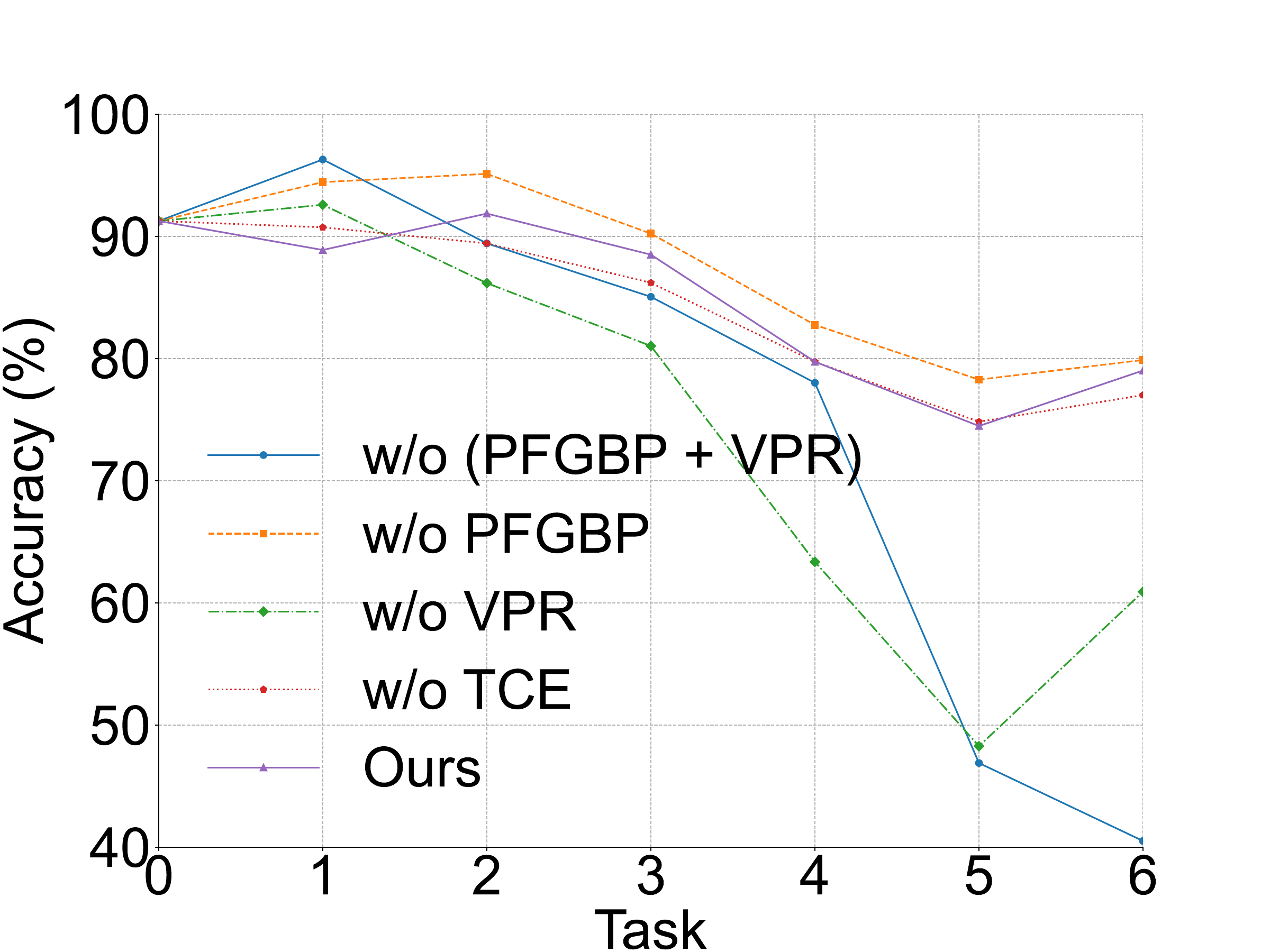}
		\label{fig:sub1}
	}
	\subfigure[Accuracy of old classes]{
		\includegraphics[width=0.22\textwidth]{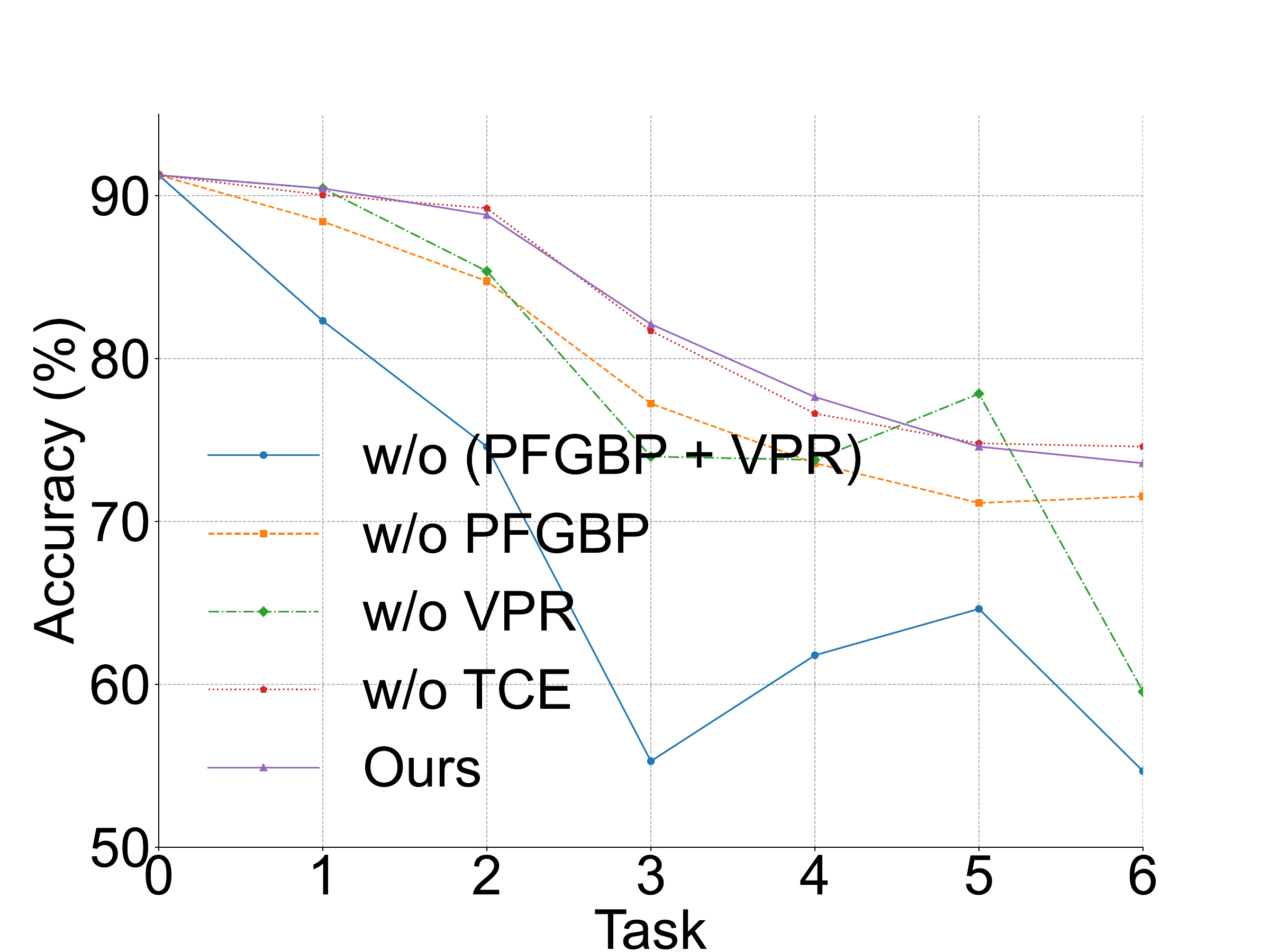}
		\label{fig:sub2}
	}
	\caption{Comparison of accuracy of new and old classes among ablated approaches for each task on the SHREC 2017 3D dataset.}
	\label{fig:new_old}
\end{figure}

\begin{figure}[htbp]
	\centering
	\begin{minipage}[t]{0.48\linewidth}
		\centering
		\includegraphics[width=\linewidth]{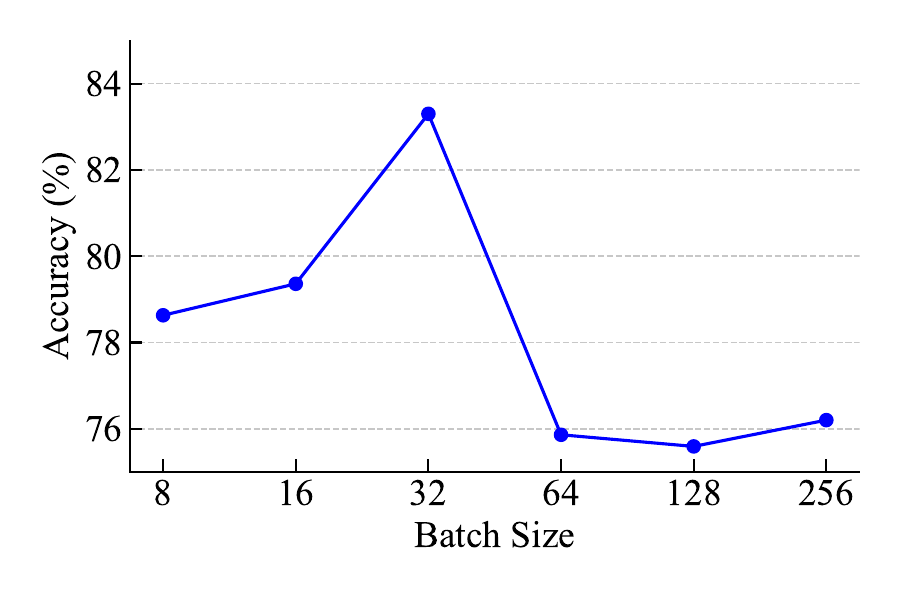}
		(a)
	\end{minipage}
	% \hspace{0.01\linewidth} 
	\begin{minipage}[t]{0.48\linewidth}
		\centering
		\includegraphics[width=\linewidth]{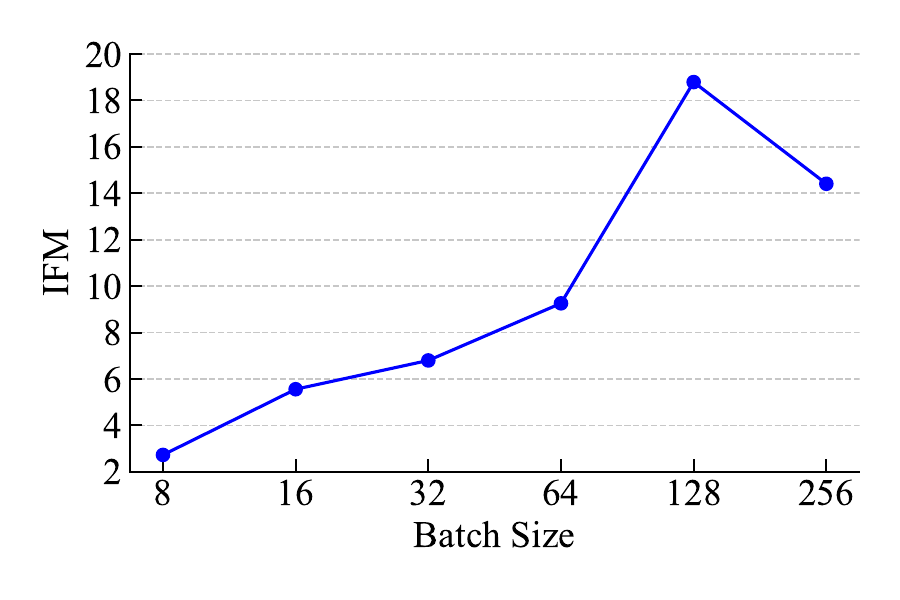}
		(b)
	\end{minipage}
	\caption{\RED{Influence of the batch size in the Prototype-Guided Pseudo Feature Replay. We report the mean global accuracies and IFM of all tasks on the SHREC 2017 3D, shown in subfigures (a) and (b), respectively.}}
	\label{fig:batch}
\end{figure}

\begin{figure*}[tbp]
	\centering
	\subfigure[Task 0 features]{
		\includegraphics[width=0.30\textwidth]{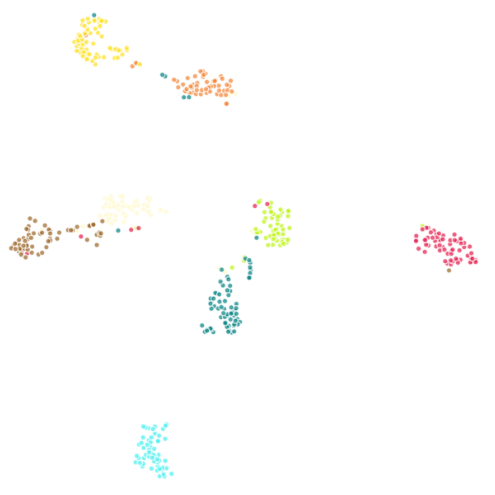}
		\label{fig:sub1}
	}
	\hfill
	\subfigure[Task 6 features (first 8 classes)]{
		\includegraphics[width=0.30\textwidth]{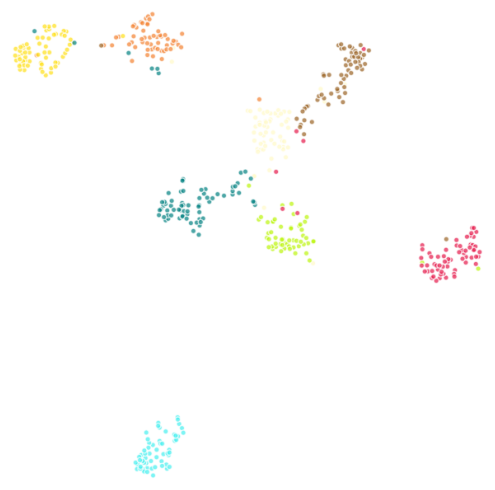}
		\label{fig:sub2}
	}
	\hfill
	\subfigure[Task 6 features (all 14 classes)]{
		\includegraphics[width=0.30\textwidth]{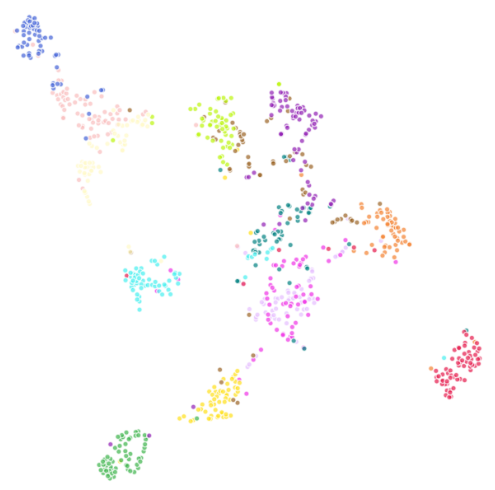}
		\label{fig:sub3}
	}
	\caption{T-SNE visualization of the deep features extracted from the SHREC 2017 3D dataset. The experiments of class-incremental learning are conducted with 7 tasks. Data points are colored according to their corresponding ground-truth labels.}
	\label{fig:fea_vis}
\end{figure*}

\begin{figure*}[t]
	\centering
	\includegraphics[width=1.0\textwidth]{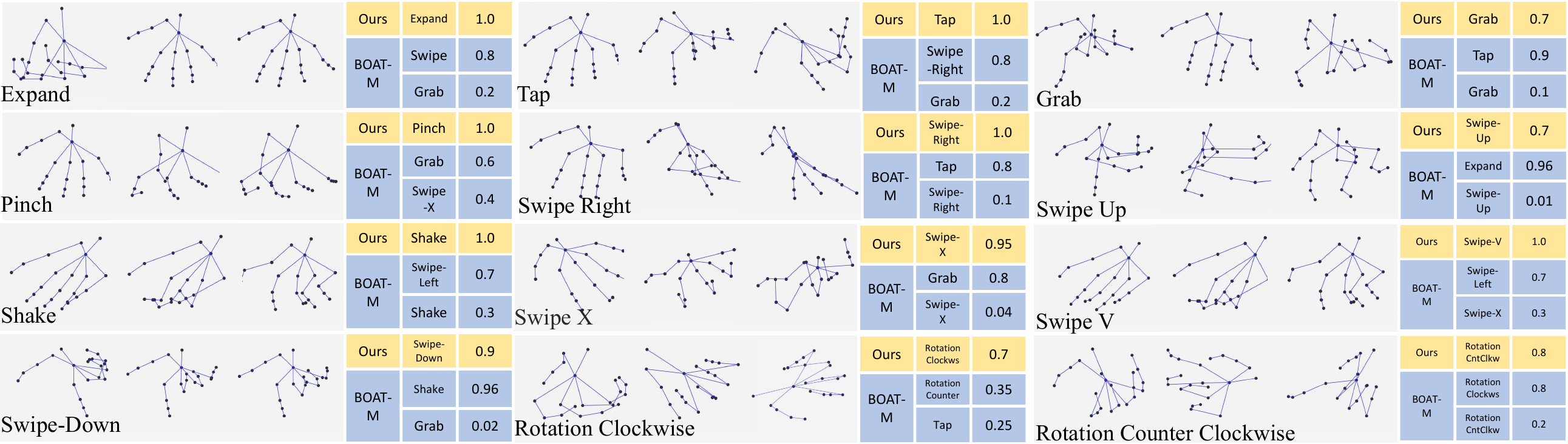}
	\caption{Visualization of gesture samples and corresponding classification results between the proposed method and BOAT-MI~\cite{aich2023data}. For a given sample, the gesture skeleton sequence is represented on the left, while the confidence scores predicted by different methods are displayed on the right.} 
	\label{fig:visualization}
\end{figure*}

% these experiments are conducted with 7 tasks. (b): features of first 8 classes from PGPFR after training on task 0, (b): features of first 8 classes from PGPFR after training on task 6, (c): features of all 14 classes from PGPFR after training on task 6. Data points are colored by their corresponding ground-truth labels. 

\begin{figure}[htbp]
	\centering
	\begin{minipage}[t]{0.48\linewidth}
		\centering
		\includegraphics[width=\linewidth]{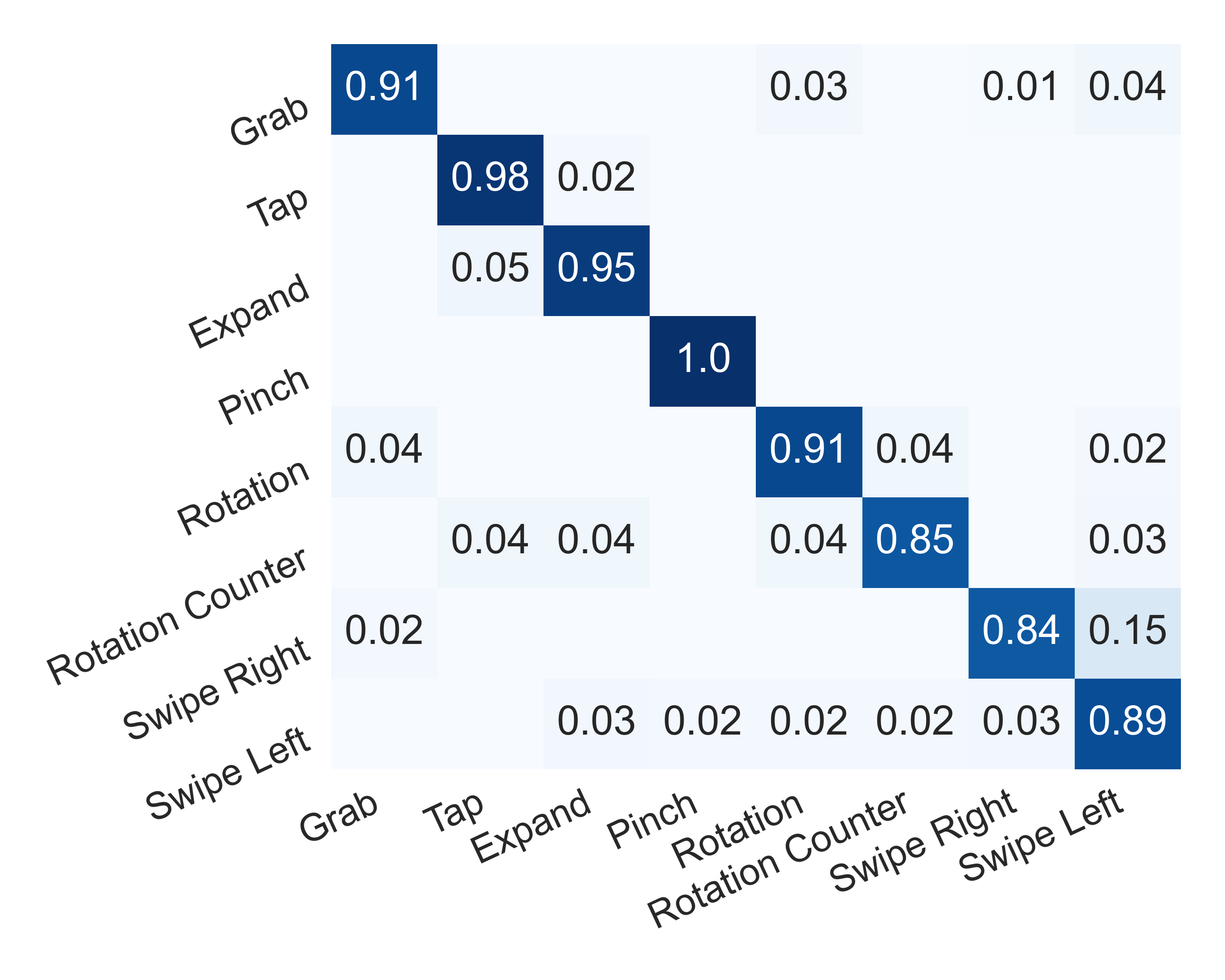}
		(a)
	\end{minipage}
	% \hspace{0.01\linewidth} 
	\begin{minipage}[t]{0.48\linewidth}
		\centering
		\includegraphics[width=\linewidth]{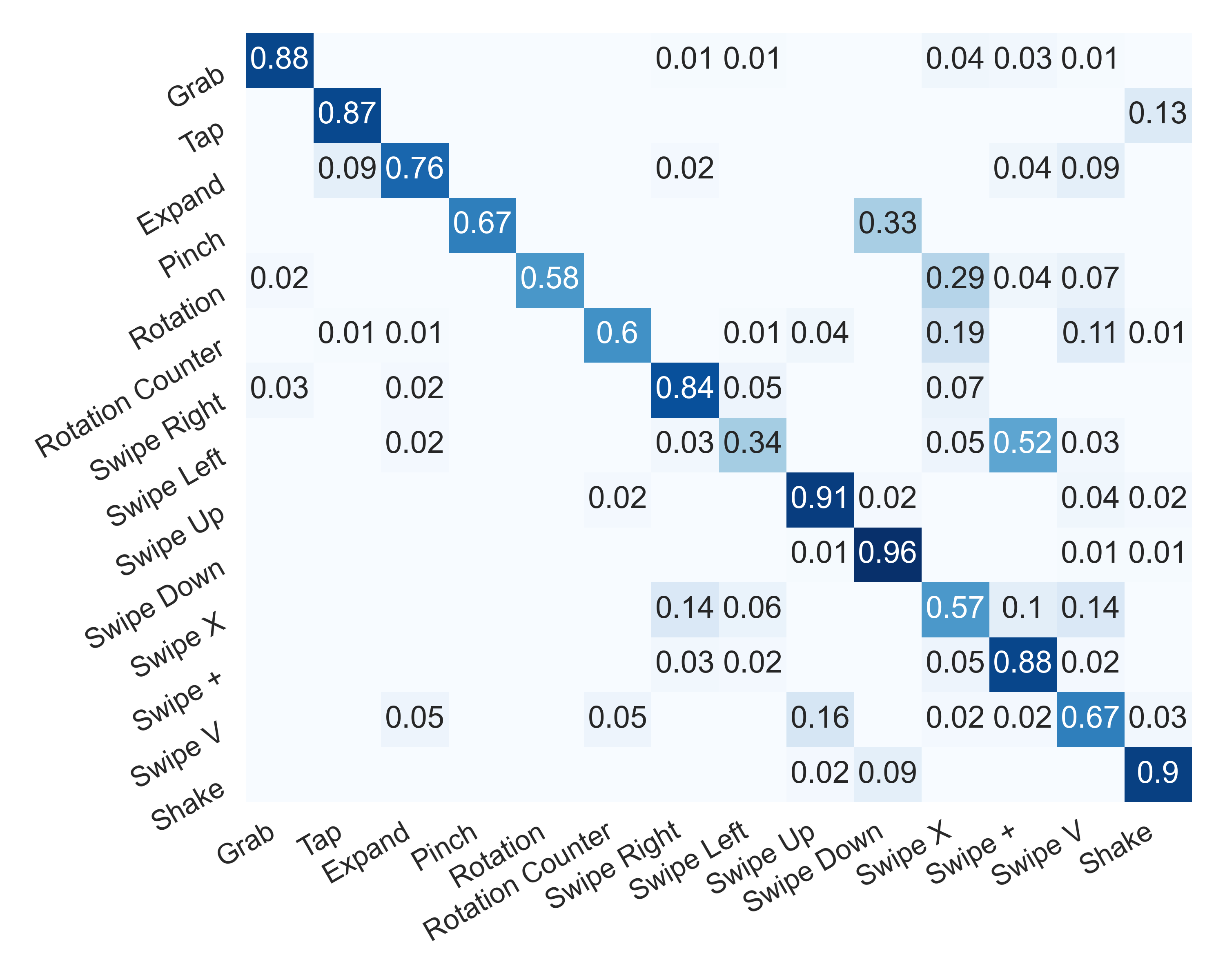}
		(b)
	\end{minipage}
	\caption{\RED{Comparison of confusion matrices before and after class-incremental gesture recognition: (a) Confusion matrix after training the Task 0. (b) Confusion matrix after training all tasks. The horizontal axis represents the predicted labels, while the vertical axis represents the ground-truth labels. The color intensity reflects the predicted probability.}}
	% , with darker shades indicating higher probabilities
	\label{fig:con_mat}
\end{figure}

\begin{table}[t]
	\caption{\RED{Comparison of training computational cost and inference speed. The evaluation platform is an Intel(R) Xeon(R) Gold 5218R CPU with an NVIDIA RTX 4090 GPU. The inversion time is calculated for the Task 2 while training on the SHREC 2017.
	}}
	\resizebox{\linewidth}{!}{
		\begin{tabular}{l|cc|cc}
			\toprule
			\multirow{2}{*}{Method} & \multicolumn{2}{c|}{Training Time (s)}    & \multicolumn{2}{c}{Inference Speed (FPS)}   \\ 
			\cmidrule{2-5}
			& Total    & Inversion            &  CPU         & GPU  \\
			\midrule
			BOAT-MI \cite{aich2023data} & 415260 & 26363 & 15  & 14 \\
			PGPFR (ours) & 1680 & 103  & 15  & 14 \\
			\bottomrule
		\end{tabular}
	}
	\label{tab:speed}
\end{table}

\subsection{Ablation Studies and Analyses}

The proposed PGPFR mainly comprises Pseudo Feature Generation with Batch Prototypes (PFGBP), Variational Prototype Replay  for Old Classes (VPR), Truncated Cross-entropy for New Classes (TCE), and Classifier Re-Training (CCRT). The PFGBP module utilizes temperature sharpening in \myref{con:eq4} and batch class prototypes in \myref{con:eq1}. The VPR module utilizes a penalty term in \myref{con:eq6}. Without this term, the regularization degenerates to PGT, as defined in \myref{con:eq5}. Detailed results of ablation studies are shown in Table \ref{tab:ablation}. 

\subsubsection{Effect of PFGBP, Batch Class Prototypes and Sharpening}
As an important component of the proposed PGPFR, PFGBP plays a crucial role in the performance of the model. Without PFGBP, we notice a decline of 1.5\% in terms of the mean Global Accuracy, while mean IFM shows a decline of 0.6. The batch class prototypes improves the global accuracy by 1.9\% as it produces more diversified pseudo features. The temperature sharpening decreases the mean IFM by 2.6, indicating that it effectively prevents catastrophic forgetting by regulating the confidences of predicted probabilities associated with pseudo features. 

\subsubsection{Effect of VPR and Prototype Replay}
Another important component is VPR, which is defined as $\mathcal{L}_{V}$. By eliminating the VPR, we observe a decline of 6.1\% in terms of the mean Global Accuracy and an increase of 0.2 in terms of the mean IFM. It proves that VPR contributes more effectively to the model’s classification ability. By comparing the performance of w/o $\mathcal{L}_{V}$ to that of w/o $\mathcal{L}_{V}$ w/ $\mathcal{L}_{Proto}$, We observe that using only the original prototype classification loss results in minimal impact on the model's performance, thereby further proving the effectiveness of the penalty term based on covariance matrices and weight vectors of the classifier. 

\subsubsection{Effect of TCE}
The TCE loss, defined as $\mathcal{L}_{T}$, encourages the model to learn the semantic information of different classes rather than emphasizing the domain discrepancy between real data and pseudo features. Removing TCE results in inferior performance for both the mean Global Accuracy and IFM.

\subsubsection{Effect of CCRT}
% The CCRT strategy provides generic and stable feature representations for subsequent incremental tasks, avoiding overfitting caused by small sample sizes or different data distributions when fine-tuning on new tasks. Without the CCRT strategy for the baseline, we observe a significant drop in the global accuracy, indicating that the representation of new classes gradually overrides the representation of old classes, leading to overfitting on the new classes.

we removed the CCRT strategy for baseline, meaning that the parameters of the feature extractor was no longer fixed. The results showed a significant drop in global accuracy after removing the CCRT strategy. This indicates that without CCRT, fine-tuning the model on new tasks causes the feature extractor to overly adapt to the new class samples, gradually overshadowing the representations of the old classes, leading to overfitting on the new classes. The CCRT is crucial for providing a general and stable feature representation, effectively preventing overfitting caused by small sample sizes or different data distributions.

\subsubsection{Further Analysis of the Accuracies of New and Old Classes}
By integrating these components, we achieve the highest Global Accuracy while maintaining a nearly minimal value of IFM. The results clearly demonstrate the effectiveness of these components in class-incremental gesture recognition.

We further analyze the roles of the components in Figure~\ref{fig:new_old}. Given that the accuracy for all visible classes and its disparity with the accuracy of the current classes in $\mathcal{D}^t$ are reported in Table~\ref{tab:ablation}, we exhibit \textit{New Accuracy} and \textit{Old Accuracy}. \textit{New Accuracy} denotes the accuracy of all visible classes excluding those from $\mathcal{D}^0$, while \textit{Old Accuracy} denotes the accuracy of classes in the initial task $\mathcal{D}^0$. We find that, without the PFGBP and the VPR, the \textit{Old Accuracy} decreases significantly during class-incremental learning, which further validates the roles of the proposed PFGBP and VPR in preventing catastrophic forgetting. Without the TCE, the \textit{Old Accuracy} remains almost unchanged, as this module is applied to new classes $\mathcal{D}^t$.

\subsubsection{\RED{Effect of Batch Size in Batch Class Prototypes}}
\RED{In Sec.~\ref{sec:batch_prototype}, we leverage batch class prototypes during online feature generation, and the default batch size is 32. Figure~\ref{fig:batch} shows the effect of batch size on performance when using batch class prototypes. We observe that the average performance first increases and then decreases as the batch size grows from 8 to 256. Although batch class prototypes improve the diversity of pseudo features for each class, a batch size that is too small prevents the prototype from adequately representing the class distribution, while an excessively large batch reduces diversity and consequently degrades performance.
}

\subsubsection{Visualizations} 
We utilize t-SNE to reduce the dimensionality of the learned features, allowing us to visualize them and qualitatively assess the effectiveness of our proposed method. The deep features $h = \varphi(f(x))$ are projected using t-SNE and are color-coded according to their ground-truth labels in the resulting visualization. By comparing Figure \ref{fig:fea_vis}(a) and Figure \ref{fig:fea_vis}(b), we observe that the first 8 classes learned from Task 0 maintain clear and distinct boundaries in the test set for Task 0. Impressively, these boundaries remain well-defined even after six incremental tasks, highlighting the strong effectiveness of our method in mitigating catastrophic forgetting. Figure \ref{fig:fea_vis}(c) displays the features of all 14 classes after learned on Task 6, using PGPFR. Recall SHREC 2017 3D is divided into 7 tasks, the visualization reveals a clear distinction between different classes.

We compare our method with the BOAT-MI \cite{aich2023data} for hand gesture recognition of certain samples in Figure \ref{fig:visualization}. Our method outperforms BOAT-MI in recognizing certain challenging samples. For instance, for the sample labeled \textit{Shake}, our model accurately identifies it with a high confidence of 1.0, while BOAT-MI mistakenly classifies it as \textit{Swipe Left} with a confidence of 0.7. Similarly, for the sample labeled \textit{Swipe Up}, despite relatively lower confidence, our method correctly classifies it, whereas BOAT-MI wrongly predicts it as \textit{Expand}. These results demonstrate the effectiveness of our method on these challenging gesture samples.

\subsubsection{\RED{Failure Case Analyses}} 
\RED{To analyze failure cases of our approach for gesture recognition. we compare the confusion matrices before and after class-incremental gesture recognition Figure \ref{fig:con_mat}. We find that the eight initial gesture categories can be correctly recognized with accuracies above 84\%. After class-incremental learning with six new categories, three of the initial categories maintain high accuracy, while the remaining five experience drops of more than 30\%. The most significant degradation occurs for swipe left, whose accuracy decreases from 89\% to 34\%, as most of its samples are misclassified as swipe V, a newly introduced category in Task 5. }

\RED{Four of the new gesture categories achieve high accuracy, while only two new categories, namely swipe X and swipe V, show relatively low accuracies of 57\% and 67\%, respectively. This reduction occurs because swipe X is confused with swipe right at 14\%, and swipe V is confused with swipe up at 16\%. These results demonstrate the effectiveness of our approach for both new and old categories in class-incremental learning.
}
% \RED{To analyze failure cases of our approach for gesture recognition, we plot the confusion matrices of the model on the test sets of Task 0 and Task 6, as shown in Figures \ref{fig:con_mat}. We observe that the diagonal values in the confusion matrix are quite high, indicating that our method effectively facilitates the model's ability to learn new classes. We also find that while maintaining an average accuracy of at least 69\% on average for the first 8 classes, the new classes also achieve high accuracy. This demonstrates that our method is crucial for preserving knowledge of the old classes. However, there are some limitations. For instance, the categories \textit{swipe up} and \textit{swipe +} exhibit similar prediction probabilities, which is likely due to the similarity between the prototypes of these two gestures. This implies that our approach might be particularly sensitive to cases where class prototypes are very similar.}

% \textbf{Complexity Analysis.} As shown in the supplementary material, our method does not necessitate the generation of pseudo-samples, and instead utilizes simple linear operations to obtain pseudo-features. Consequently, it exhibits exceptional efficiency, with a time complexity of $O(n)$. From the perspective of required memory, this method only needs to store the class prototype and covariance matrix. Furthermore, compared to the pseudo samples generated by BOAT-MI\cite{aich2023data}, which need to be stored for subsequent tasks, pseudo features generated by our method can be discard after use, resulting a low space complexity.

\subsubsection{\RED{Training and Inference Efficiency Analysis}}
\RED{We report the training time and inference speed of our approach in Table~\ref{tab:speed}. Our method only introduces additional computation during training, while the inference network remains identical to that of BOAT-MI \cite{aich2023data}. The inference speed reaches approximately 15 FPS on both CPU and GPU platforms. However, our training time is only about 0.4\% of that required by BOAT-MI.}

\subsubsection{Limitations}
\RED{One limitation is that our pseudo feature generation relies on simple linear operations. Although this design improves efficiency, it cannot accurately approximate the true distributions of old categories. Consequently, when a new category is highly similar to an existing one, our model suffers catastrophic forgetting on that old category.} 

\section{Conclusion}
In this work, we study class-incremental gesture recognition and introduce a Prototype-Guided Pseudo Feature Replay (PGPFR). This approach primarily consists of Pseudo Feature Generation with Batch Prototypes (PFGBP), Variational Prototype Replay, Truncated Cross-Entropy and Continual Classifier Re-Training. Extensive experiments on popular gesture recognition benchmarks demonstrate the effectiveness of the proposed PGPFR as well as the components. Specifically, PFGBP and Variational Prototype Replay collaboratively tackle the issue of catastrophic forgetting of old knowledge when learning new classes. Truncated Cross-Entropy mitigates the domain difference between real data and pseudo features and improves performance for new classes. We also discover that classifier re-training enhances the performance of class-incremental learning by potentially preventing overfitting to instances of new classes. We believe that it will considerably contribute to research in both data-free class-incremental learning and gesture recognition.

% \fi
\IEEEpeerreviewmaketitle

\ifCLASSOPTIONcaptionsoff
  \newpage
\fi

\bibliographystyle{IEEEtran}
\bibliography{IEEEtran}

\end{document}